\documentclass[sigconf]{acmart}
\usepackage{algorithm}
\usepackage{algpseudocode}
\usepackage{tabu}
\usepackage{array}
\AtBeginDocument{%
  }

\copyrightyear{2025} 
\acmYear{2025} 
\setcopyright{acmlicensed}\acmConference[WWW Companion '25]{Companion Proceedings of the ACM Web Conference 2025}{April 28-May 2, 2025}{Sydney, NSW, Australia}
\acmBooktitle{Companion Proceedings of the ACM Web Conference 2025 (WWW Companion '25), April 28-May 2, 2025, Sydney, NSW, Australia}
\acmDOI{10.1145/3701716.3715211}
\acmISBN{979-8-4007-1331-6/2025/04}

\begin{document}

\title{Balancing Efficiency and Effectiveness: An LLM-Infused Approach for Optimized CTR Prediction}

\author{Guoxiao Zhang}
\authornote{Both authors contributed equally to this research.}
\email{zhangguoxiao@meituan.com}
\orcid{0009-0004-2628-7094}
\author{Yi Wei}
\authornotemark[1]
\email{weiyi20@meituan.com}
\affiliation{%
  \institution{Meituan}
  \city{Beijing}
  \country{China}
}
\author{Yadong Zhang}
\affiliation{%
  \institution{Meituan}
  \city{Beijing}
  \country{China}}
\email{zhangyadong05@meituan.com}

\author{Huajian Feng}
\affiliation{%
  \institution{Hunan University}
  \city{Changsha}
  \country{China}}
\email{huajianfeng@hnu.edu.cn}

\author{Qiang Liu}
\affiliation{%
  \institution{Meituan}
  \city{Beijing}
  \country{China}}
  \email{liuqiang43@meituan.com}

\renewcommand{\shortauthors}{Guoxiao Zhang, Yi Wei, Yadong Zhang, Huajian Feng, and Qiang Liu}

\begin{abstract} 
Click-Through Rate (CTR) prediction is essential in online advertising, where semantic information plays a pivotal role in shaping user decisions and enhancing CTR effectiveness. Capturing and modeling deep semantic information, such as a user's preference for "Häagen-Dazs' HEAVEN strawberry light ice cream" due to its health-conscious and premium attributes, is challenging. Traditional semantic modeling often overlooks these intricate details at the user and item levels. To bridge this gap, we introduce a novel approach that models deep semantic information end-to-end, leveraging the comprehensive world knowledge capabilities of Large Language Models (LLMs). Our proposed LLM-infused CTR prediction framework(\textbf{M}ulti-level Deep \textbf{S}emantic Information Infused CTR model via \textbf{D}istillation, MSD) is designed to uncover deep semantic insights by utilizing LLMs to extract and distill critical information into a smaller, more efficient model, enabling seamless end-to-end training and inference. Importantly, our framework is carefully designed to balance efficiency and effectiveness, ensuring that the model not only achieves high performance but also operates with optimal resource utilization. Online A/B tests conducted on the Meituan sponsored-search system demonstrate that our method significantly outperforms baseline models in terms of Cost Per Mile (CPM) and CTR, validating its effectiveness, scalability, and balanced approach in real-world applications.
\end{abstract}



\maketitle

\section{Introduction}
\begin{figure}[ht]
    \centering
    \includegraphics[width=\linewidth]{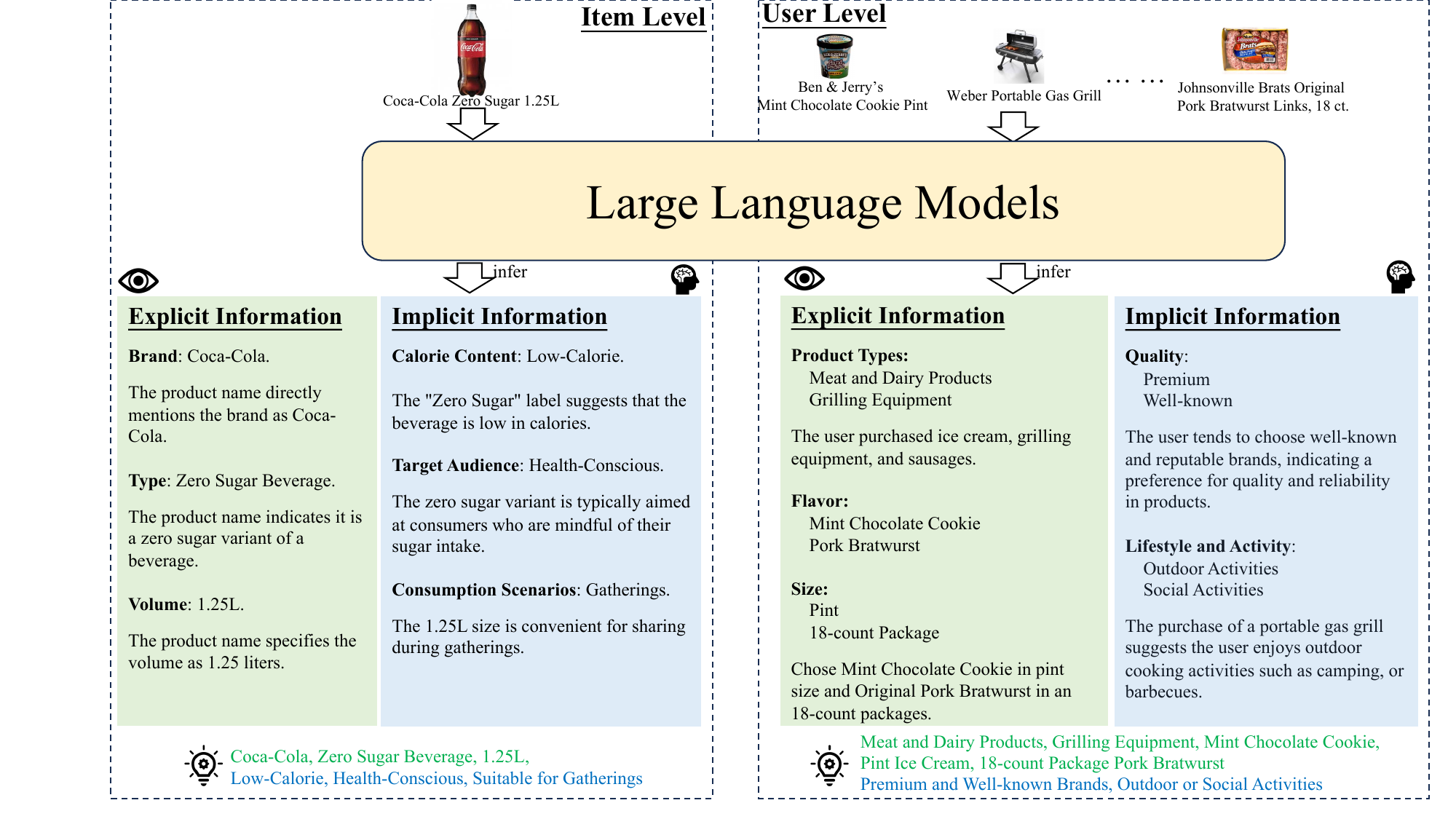}
    \caption{Semantic information at both item and user levels: explicit and implicit key features with corresponding reasoning processes extracted by LLMs.}
    \label{fig:infer}
\end{figure}

Click-Through Rate (CTR) prediction plays an important role in recommender systems and online advertising \cite{guo2017deepfm,zhou2018deep,gligorijevic2019deeply,bian2020can,pi2020search,wang2022learning,hong2024print}. Lately, some works are proposed to capture semantic information by involving Pretrained Language Models (PLMs) \cite{wang2022learning,wang2023bert4ctr,yang2023practice,li2023ctrl,lin2024clickprompt,xi2024towards,geng2024breaking}. As is shown in Figure \ref{fig:infer}, semantic information can be categorized as explicit information and implicit information, where explicit information is directly obtainable from the texture features and implicit information is inferred by relevant world knowledge. 

Compared with those works of BERT series \cite{wang2022learning,wang2023bert4ctr,yang2023practice}, generative Large Language Models (LLMs) provide a fresh technological approach for recommender systems to capture deep semantic information\cite{li2023ctrl,lin2024clickprompt,xi2024towards,geng2024breaking}, due to their capacity for leveraging world knowledge to infer implicit information. Specifically, CTRL\cite{li2023ctrl} integrates semantic information from LLMs into traditional ID-based models using contrastive learning. ClickPrompt \cite{lin2024clickprompt} first aligns the collaborative and semantic knowledge from ID and textual features via the prompt interface, then tunes the CTR model without PLM for inference efficiency. KAR \cite{xi2024towards} acquires two types of external knowledge from LLMs—the reasoning knowledge on user preferences and the factual knowledge on items, which are transformed into augmented vectors to be compatible with the recommendation task. BAHE\cite{geng2024breaking} employs the LLM’s pre-trained low layers to extract embeddings of user behaviors and stores them in the offline database, then utilizes the deeper, trainable layers of the LLM to generate comprehensive user embeddings. However, it is important to note that, due to limited consideration of efficiency, most of these studies(ClickPrompt\cite{lin2024clickprompt}, KAR \cite{xi2024towards}, CTRL\cite{li2023ctrl}) remain largely experimental and have not been applied in practical industrial settings for CTR prediction. Besides, BAHE\cite{geng2024breaking} chooses to partially utilize information from LLMs, which restricts the exploration of the effectiveness of LLMs for CTR estimation.

To balance efficiency and effectiveness, inspired by the fact Knowledge distillation is an effective model compression method that can transfer the internal capabilities of LLMs to smaller ones\cite{gu2024minillm},  we propose an LLM-infused approach via Knowledge Distillation. Specifically, our approach includes the Multi-level Knowledge Distillation Module(MKDM) and Multi-level Knowledge Integration Module(MKIM). For MKDM, we propose a prompt-based approach combined with Chain-of-Thought (CoT) reasoning to generate reasoning processes and extract both explicit and implicit information at both item and user levels, shown in Figure \ref{fig:infer}. The MIDM comprises three components. Initially, we incorporate a feature adaptation layer to map the semantic embeddings into the appropriate feature space. Subsequently, we employ LoRA (Low-Rank Adaptation) to fine-tune the parameters of the LLM. Finally, we develop a Frequency-Adaptive k-Near Items Fusion to enhance integration with the CTR model.

Our contributions include: 
\begin{itemize}
\item Introducing an LLM-infused CTR Prediction Approach through Knowledge Distillation Paradigm to balance efficiency and effectiveness. 
\item Developing Multi-level Knowledge Distillation Module and Multi-level Knowledge Integration Module. 
\item Successfully deploying this method on the Meituan Recommendation platform, achieving a 2.12\% rise in CTR and a 2.59\% enhancement in Cost Per Mile (CPM) during online A/B tests.
\end{itemize}

\section{Methodology}

\begin{figure}[ht]
    \centering
    \includegraphics[width=\linewidth]{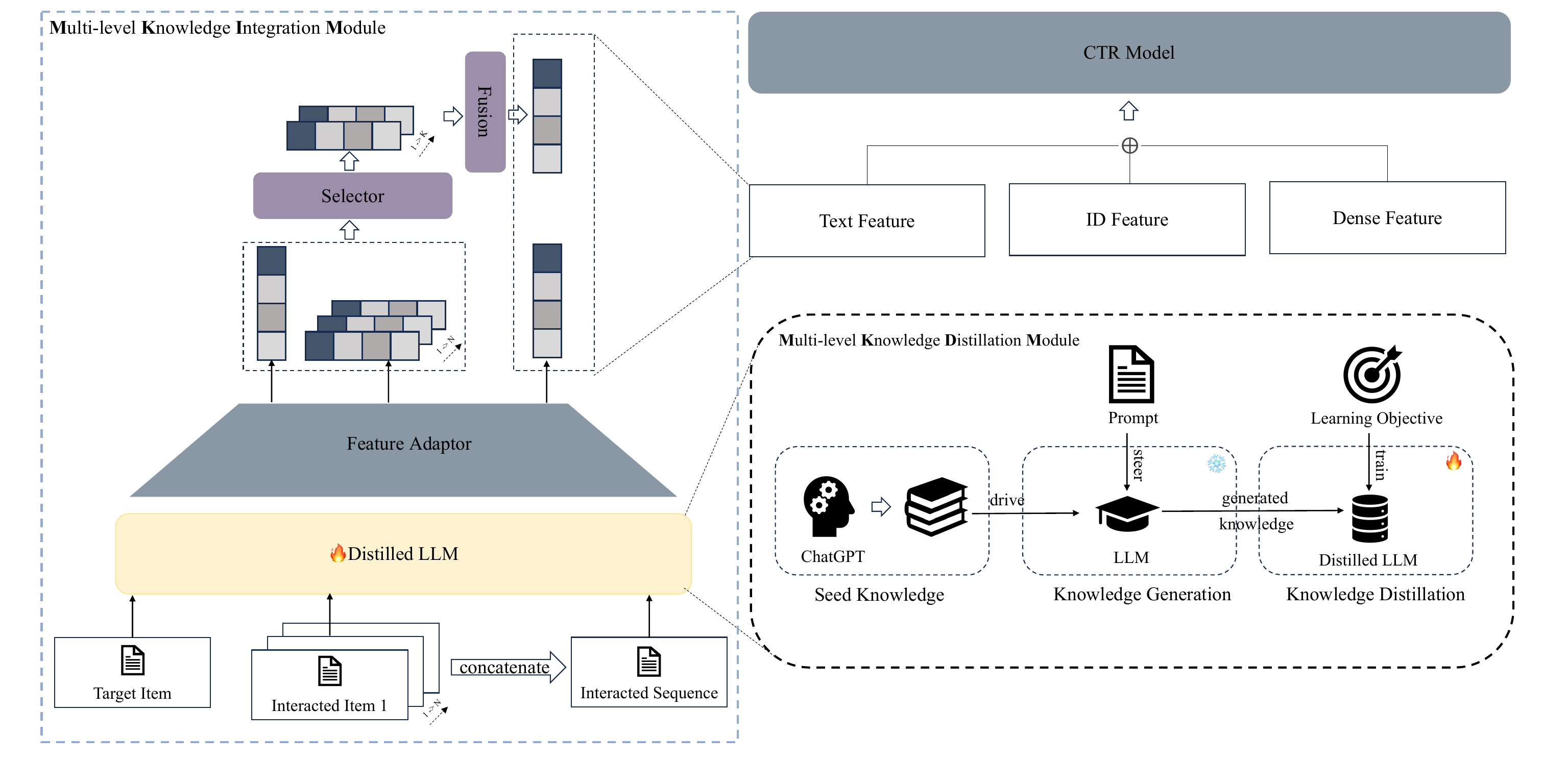}
    \caption{The overall framework of our proposed LLM-infused CTR prediction framework(MSD).}
    \label{fig:framework}
\end{figure}

In this section, we present our framework, depicted in Figure \ref{fig:framework}, which consists of two main modules: the Multi-level Knowledge Distillation Module and the Multi-level Knowledge Integration Module. We first introduce the process of distilling LLMs at both item and user levels, followed by integrating the distilled model into the CTR prediction framework.

\subsection{Multi-level Knowledge Distillation Module}

This module is designed to efficiently distill semantic knowledge at both item and user levels from LLMs into a more efficient one. Knowledge generation and knowledge distillation are the main components of this module. We provide a detailed description of them in this section.

\subsubsection{Knowledge Generation}

\begin{figure}[ht]
    \centering
    \includegraphics[width=\linewidth]{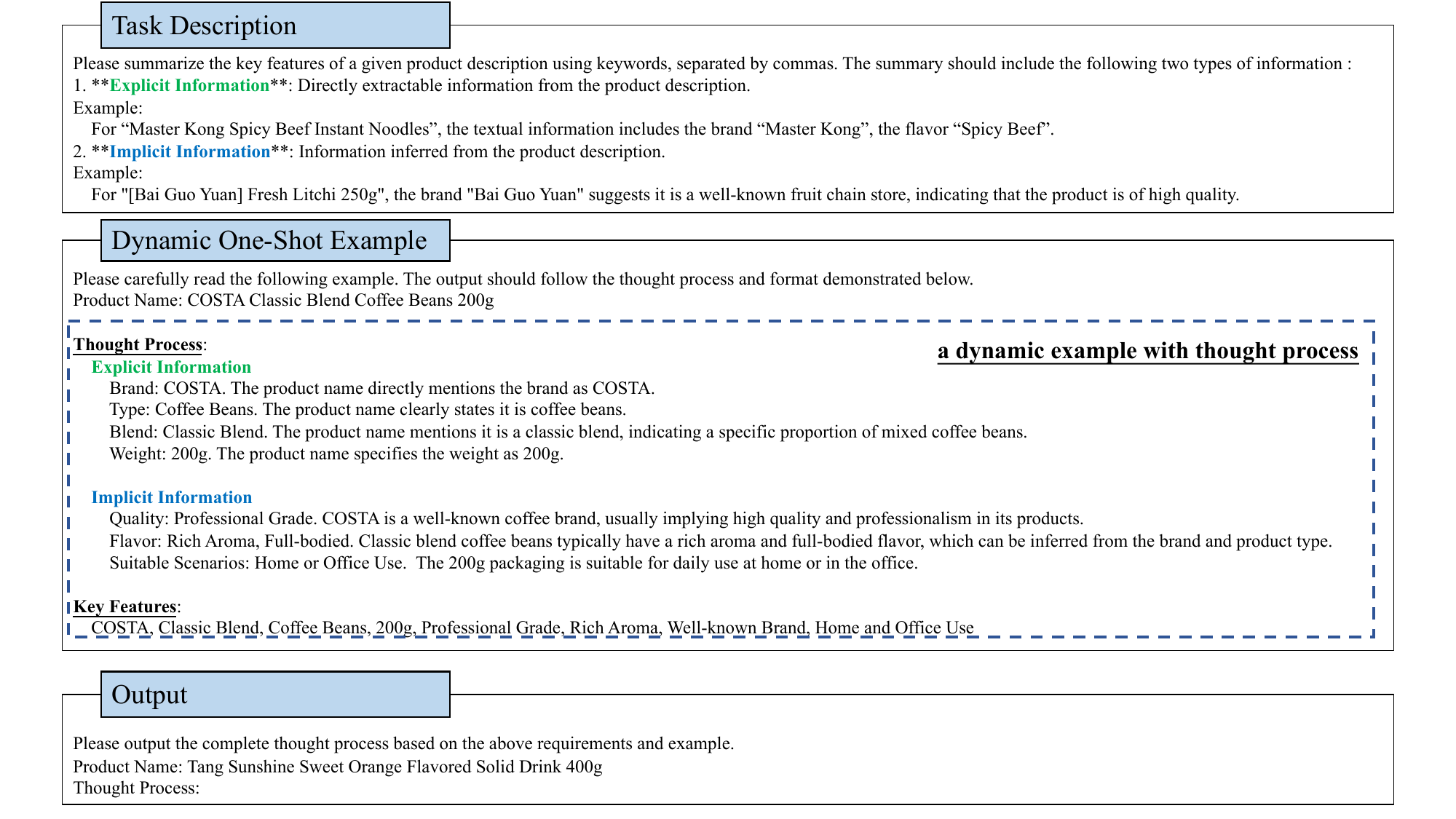}
    \caption{Item Level Prompt templates guide the LLM in generating outputs containing explicit (highlighted in green) and implicit (highlighted in blue) information, utilizing in-context learning.}
    \label{fig:cot_few_shot_prompt}
\end{figure}

High-quality and diverse data are essential to ensure that the distilled model generalizes well across different contexts and retains critical semantic information. To guide the LLMs produce high-quality data, we first manually selected outputs from the ChatGPT as reference, which include key phrases and the rationale. Then, the prompt will be enhanced by dynamically chosen reference output as an example for in-context learning\cite{d2024dynamic}. Our prompt template for item level is illustrated in Figure \ref{fig:cot_few_shot_prompt}. To ensure the dataset is representative and diverse, stratified and importance sampling are adopted, considering categorical information and posterior statistics. For user level data, the generation procedure follows a similar protocol.

\subsubsection{Knowledge Distillation}

Following \cite{wang2024can}, we leverage knowledge distillation to transfer semantic and contextual reasoning capabilities from the LLMs teacher model \( T \) to a more computationally efficient student model \( S \). The student model is trained to replicate the output \( y_T \) of the teacher model by minimizing the distillation loss as follows:
\begin{equation}
\mathcal{L}_{\text{distill}} = \sum_{t=1}^{|\mathbf{y}_T|} \log \left( P_{\theta} \left( \mathbf{y}_{T,t} \mid \mathbf{y}_{S,<t}, \mathbf{x} \right) \right),
\end{equation}
\begin{equation}
S^* = \arg \min_{S} \mathcal{L}_{\text{distill}}
\end{equation}
where \( \mathbf{y}_{T,t} \) represents the \( t \)-th token of the teacher model's output sequence, \( \mathbf{y}_{S,<t} \) are the preceding tokens in the student model's output sequence and \(\mathbf{x}\) are the input tokens.

\subsection{Multi-level Knowledge Integration Module} 
The MKIM is designed to seamlessly integrate semantic insights at both the item and user levels within the CTR prediction framework. It comprises three key components: 1) LoRA, which enhances alignment and performance by fine-tuning distilled LLMs with Low-Rank Adaptors for computational efficiency; 2) Feature Adaptors, which compress and transform LLM output embeddings using a Multi-Layer Perceptron (MLP) to ensure effective integration into the CTR model; 3) Frequency-Adaptive Relevant Items Fusion, which improves item embedding robustness through masking and pooling operations. The transformed item and user embeddings are then concatenated to form the input for CTR tasks.

\subsubsection{LoRA} 
Following \cite{hu2021lora}, we finetune the distilled LLMs for CTR tasks using Low-Rank adaptors. Specifically, for a pre-trained weight matrix $\mathbf{W}_0 \in \mathbb{R}^{d\times k}$, we constrain its update by representing the latter with a low-rank decomposition $\mathbf{W_0} + \Delta \mathbf{W} = \mathbf{B} \mathbf{A}$, where $\mathbf{B} \in \mathbb{R}^{d\times r}$, $\mathbf{A} \in \mathbb{R}^{r\times k}$, and the rank $r \ll \min(d, k)$. During training, $\mathbf{W}_0$ is frozen and does not receive gradients, while the $\mathbf{A}$ and $\mathbf{B}$ contain trainable parameters. 

\subsubsection{Feature Adaptors} 

For the raw output embeddings of the LLMs, we define \(\mathbf{e}_{\text{u}}\) as the user-level embedding, \(L \) as the length of user behavior sequence, \(\mathbf{e}_{\text{t}}\) as the target item embedding and \(\mathbf{e}_{\text{i}}\) as the item embeddings in user behavior sequence, where \(i\) ranges from 1 to \(L\). We apply a Multi-Layer Perceptron (MLP) as the feature adaptor to generate the projected embeddings \(\mathbf{e}^{\prime}_{\text{u}}\), \(\mathbf{e}^{\prime}_{\text{t}}\) and \(\mathbf{e}^{\prime}_{\text{i}}\) as follows:
\begin{equation}
\mathbf{e}^{\prime}_{\text{u}} = \text{MLP}(\mathbf{e}_{\text{u}})
\end{equation}
\begin{equation}
\mathbf{e}^{\prime}_{\text{t}} = \text{MLP}(\mathbf{e}_{\text{t}})
\end{equation}
\begin{equation}
\mathbf{e}^{\prime}_{\text{i}} = \text{MLP}(\mathbf{e}_{\text{i}}), \quad i \in \{1, \dots, L\}
\end{equation}

\subsubsection{Frequency-Adaptive Relevant Items Fusion} 
To enhance the robustness of item embeddings during training, the item-level features \(\mathbf{e}_{\text{item}}\) are obtained as follows:
\begin{equation} 
\text{sim}(\mathbf{e}^{\prime}_{t}, \mathbf{e}^{\prime}_{i}) = \frac{\mathbf{e}^{\prime}_{t} \cdot \mathbf{e}^{\prime}_{\text{masked\_i}}}{\|\mathbf{e}^{\prime}_{t}\| \|\mathbf{e}^{\prime}_{\text{masked\_i}}\|}
\end{equation}

\begin{equation} 
\mathbf{e}_{\text{top-k}} = \text{Top-k}\left\{\text{sim}(\mathbf{e}^{\prime}_{t}, \mathbf{e}^{\prime}_{i}) \mid i \in \{1, \dots, L\}\right\}
\end{equation}

\begin{equation} 
\mathbf{e}_{\text{item}} = \sum_{i=1}^{k} \mathbf{e}_{\text{top-k}, i}
\end{equation}
where \( \mathbf{e^{\prime}_{\text{t}}} \) represents the target item, \( \mathbf{e^{\prime}_{\text{masked\_i}}} \) is gained from  \( \mathbf{e^{\prime}_{\text{i}}} \) with frequency-guided stochastic masking. The top-k similar item embeddings are then processed through a fusion layer to produce the final item embedding. We utilize the element-wise addition as our fusion operation.

\section{EXPERIMENTS}

\subsection{Experimental Setup}

\subsubsection{Dataset.} 
We conducted experiments using both a public dataset and one real-world business dataset. The key statistics of these datasets are presented in Table \ref{tab:dataset_statistics}.

\begin{itemize}
    \item \textbf{KDD Cup 2012}: A CTR prediction dataset comprising advertising logs from Tencent, which include queries and user information. We randomly selected 10\% of the users from the complete dataset for computation efficiency.
    \item \textbf{Meituan}: A real-world dataset, sourced from Meituan's recommendation platform, contains extensive user-item interaction data that reflects actual business scenarios.
\end{itemize}

\begin{table}[h]
\centering
\caption{Statistics of datasets used in our experiments.}
\label{tab:dataset_statistics}
\begin{tabular}{lccc}
\toprule
\textbf{Dataset} & \textbf{\#users} & \textbf{\#items} & \textbf{\#interactions} \\
\hline
KDD Cup 2012 & 2,202,355 & 288,794 & 13,624,635 \\
Meituan & 152,287,209 & 160,837,520 & 2,304,146,513 \\
\bottomrule
\end{tabular}
\end{table}

\subsubsection{Baselines.} 
To demonstrate the effectiveness of our proposed framework, we compare it with the popular CTR baseline models (parameter settings following \cite{hong2024print}):

\begin{itemize}
    \item DIN \cite{zhou2018deep}: a leading sequential CTR model that uses an attention mechanism to identify target-related interests.
    \item DeepFM \cite{guo2017deepfm}: a popular CTR model that merges factorization machines with deep learning techniques for feature interaction.
    \item DeepCharMatch \cite{gligorijevic2019deeply}: a method that directly models semantic information at the character level using query-ad pairs.
    \item SuKD \cite{wang2022learning}: a model that incorporates additional NLP knowledge in the fine-tuning of PLMs.
    \item BERT4CTR \cite{wang2023bert4ctr}: an effective framework for integrating PLMs with non-textual features.
    \item PRINT \cite{hong2024print}: a BERT-based model that improves CTR prediction by accounting for personalized incentives in query-ad semantic relevance.
\end{itemize}

\subsubsection{Evaluation Metrics.} 
To thoroughly assess our approach, we adopted distinct metrics for evaluating the effectiveness of the distillation process and the model's overall performance while examining the relationship between these two evaluations.

\paragraph{Distillation Evaluation Metric.} To evaluate the effectiveness of our distillation process, we calculate the F1 score based on predicted key phrases. Cosine similarity is applied to determine phrase equivalence, leveraging embeddings generated by a BERT model. We manually annotated a subset of the LLM's outputs to establish a ground truth.

\paragraph{Model Performance Metric.} The primary metric for evaluating the model's performance is the Area Under the ROC Curve (AUC), a widely recognized measure of a model's capability to distinguish between positive and negative instances. In addition, we utilize the Relative Improvement (RelaImpr) metric \cite{yan2014coupled} to quantify the comparative enhancement our model achieves over existing models.

\subsection{Overall Performance}
In this section, we evaluate the performance of our novel LLM-infused CTR prediction framework against PRINT\cite{hong2024print} and other baseline models across the two datasets. As detailed in Table 2, our approach demonstrates superior performance. Our framework leverages the deep semantic capabilities of Large Language Models as well as end-to-end training and inference processes. Compared to PRINT, our approach achieves a notable improvement in AUC, with increases of 0.25\% and 0.63\% over the KDD Cup 2012 and Meituan datasets, respectively.

\begin{table}[h]
    \centering
    \caption{Performance comparison of our proposed MSD.}
    \begin{tabular}{lccccc}
        \toprule
        \textbf{Models} & \multicolumn{2}{c}{\textbf{KDD Cup 2012}} & \multicolumn{2}{c}{\textbf{Meituan}} \\ 
        \cmidrule(lr){2-3} \cmidrule(lr){4-5}
        & \textbf{AUC} & \textbf{RelaImpr} & \textbf{AUC} & \textbf{RelaImpr} \\ 
        \midrule
        DeepFM & 0.7763 & 0.00\% & 0.6938 & 0.00\% \\ 
        DIN & 0.7792 & 1.05\% & 0.6963 & 1.29\% \\ 
        DeepCharMatch & 0.7794 & 1.12\% & 0.6967 & 1.50\% \\ 
        SuKD & 0.7821 & 2.10\% & 0.6960 & 1.14\% \\ 
        NumBERT & 0.7832 & 2.50\% & 0.6983 & 2.32\% \\ 
        BERT4CTR & 0.7838 & 2.71\% & 0.7004 & 3.41\% \\ 
        PRINT & 0.7846 & 3.00\% & 0.7024 & 4.43\% \\ 
        \hline
        \textbf{Our MSD} & \textbf{0.7871} & \textbf{3.91\%} & \textbf{0.7087} & \textbf{7.64\%} \\ 
        \bottomrule
    \end{tabular}
\end{table}

\subsection{Ablation Study for Our Framework}

We performed a comprehensive ablation study to evaluate the contribution of each module within our framework. As depicted in Table \ref{tab:ablation_study}, each module significantly contributes to the overall performance improvement of the complete model.

\begin{table}[h]
    \centering
    \begin{tabular}{lc}
        \toprule
        \textbf{Methods} & \textbf{AUC} \\ \hline
        w/o LoRA & 0.7075\textsuperscript{(-0.12\%)} \\
        w/o Item Level Fusion & 0.7069\textsuperscript{(-0.18\%)} \\ 
        w/o User Level & 0.7054\textsuperscript{(-0.33\%)} \\
        \textbf{MSD (our full model)} & \textbf{0.7087} \\
        \bottomrule
    \end{tabular}
    \caption{Ablation study of the MSD over the Meituan datasets.}
    \label{tab:ablation_study}
\end{table}

\subsection{Effectiveness of LLM Distillation}

\begin{figure}
    \centering
    \includegraphics[width=\linewidth]{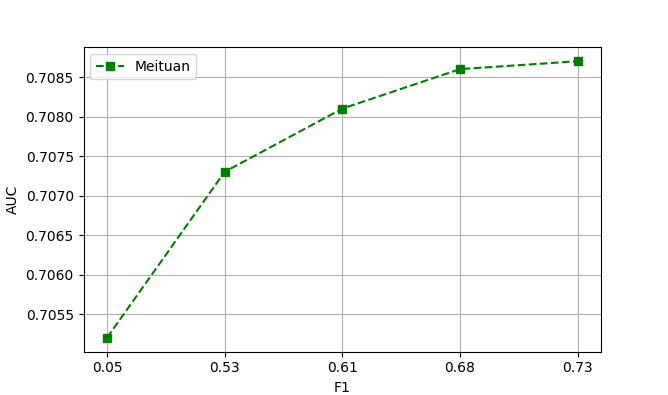}
    \caption{Relationship between distillation metric F1 score and CTR model performance metric AUC on the Meituan dataset. The initial low F1 score improves to 0.73 with additional fine-tuning data and data augmentation strategies, such as key phrase shuffling and dictionary substitutions.}
    \label{fig:f1_vs_auc}
\end{figure}

Our experiments demonstrate a positive correlation between the knowledge distillation process and the performance of the CTR model, highlighting the critical role of an effective distillation strategy in enhancing model performance. Specifically, when the F1 score increases from \( 0.05 \) to \( 0.53 \), the AUC score of the CTR model shows a significant improvement. However, beyond an F1 score of \( 0.60 \), the rate of increase in the AUC score begins to diminish as shown in Figure \ref{fig:f1_vs_auc}. This observation suggests that the current method can enhance CTR model performance only up to a certain level. To further improve the CTR model, future research could explore more sophisticated designs for the distillation task to overcome this performance plateau.

\subsection{Online Deployment and A/B Test Results}
To reduce the inference time of the CTR model, we implement a combination of pre-computation and a hierarchical caching system, as inspired by PRINT\cite{hong2024print}. For target item embeddings, we store pre-computed embeddings of top items ranked by exposure counts in Redis. If an item is not found in the cache, its embedding is computed in real time after the recall phase and before being fed into the CTR model. These newly computed embeddings are then cached in Redis, utilizing an eviction policy to manage storage efficiently.

\begin{table}[ht]
    \centering
    \caption{Results of Online A/B Testing.}
    \begin{tabular}{cccc}
        \toprule
        \textbf{Method} & \textbf{CTR Gain} & \textbf{CPM Gain} & \textbf{Inference Time} \\
        \hline
        Baseline & - & - & 35 ms \\
        \textbf{MSD} & $\textbf{+2.12\%}$ & $\textbf{+2.59\%}$ & 37.2 ms \\
        \bottomrule
    \end{tabular}
    \label{tab:ab_test_results}
\end{table}
Between October 20, 2024, and October 30, 2024, we conducted an online A/B test on Meituan's sponsored search advertising system to validate our proposed framework. As illustrated in Table \ref{tab:ab_test_results}, our model demonstrates a significant improvement over the naive implementation of the LLM, achieving a 2.12\% increase in Click-Through Rate (CTR) and a 2.59\% increase in Cost Per Mille (CPM).

\section{Conclusion}
In this paper, we present an innovative LLM-infused framework for CTR prediction designed to effectively utilize semantic information in CTR tasks. By introducing the MKDM and MKIM, our framework captures deep semantic insights at both item and user levels, addressing the limitations of traditional ID-based recommendation systems and existing methods that attempt to integrate LLMs or BERT into CTR tasks. Our model significantly enhances the AUC on the Meituan dataset and achieves a notable increase in system revenue by 2.59\%, and CTR by 2.12\% in the Meituan sponsored-search system. This approach opens new avenues for incorporating LLMs into CTR prediction, and we hope our work inspires further research and exploration in this field.

\bibliographystyle{ACM-Reference-Format}
\balance
\bibliography{sample-base}
\end{document}